\documentclass[runningheads]{llncs}

\usepackage{eccv}

\usepackage{eccvabbrv}

\usepackage{graphicx}
\usepackage{booktabs}
\usepackage{placeins}
\usepackage{tcolorbox}
\usepackage{colortbl}
\usepackage{lipsum}
\usepackage{multirow}
\usepackage{pifont}
\usepackage{wrapfig}
\usepackage{makecell}
 
\usepackage[accsupp]{axessibility}  

\usepackage[pagebackref,breaklinks,colorlinks,citecolor=eccvblue]{hyperref}

\usepackage{orcidlink}

\begin{document}

\title{7ABAW-Affective Behaviour Analysis via Progressive Learning} 

\titlerunning{Affective Behaviour Analysis via Progressive Learning}
\author{Chen Liu\inst{1,2}\textsuperscript{*} \orcidlink{0000-0003-3159-0034}\and
Wei Zhang\inst{1}\textsuperscript{*} \orcidlink{0000-0001-5907-7342}\and
Feng Qiu\inst{1} \orcidlink{0000-0002-4333-5957}\and
Lincheng Li\inst{1}\textsuperscript{$\dagger$} \orcidlink{0000-0002-6047-0472}\and
Dadong Wang\inst{3}\orcidlink{0000-0003-0409-2259} \and
Xin Yu\inst{2}\orcidlink{0000-0002-0269-5649}}

\authorrunning{C.~Liu et al.}

\institute{NetEase Fuxi AI Lab, Hangzhou, China \\
\email{\{zhangwei05, qiufeng, lilincheng\}@corp.netease.com} \and
The University of Queensland, Queensland, Australia\\ 
\email{chen.liu7@uqconnect.edu.au, xin.yu@uq.edu.au} 
\and 
CSIRO, Data61, Sydney, Australia\\
\email{Dadong.Wang@data61.csiro.au}
\footnote[1]{* Equal Contribution, $\dagger$ Corresponding authors. This work is done at Netease.}
}

 
\maketitle

\begin{abstract}
Affective Behavior Analysis aims to develop emotionally intelligent technology that can recognize and respond to human emotions. 
To advance this field, the 7th Affective Behavior Analysis in-the-wild (ABAW) competition holds the Multi-Task Learning Challenge based on the s-Aff-Wild2 database.
The participants are required to develop a framework that achieves Valence-Arousal Estimation, Expression Recognition, and AU detection simultaneously.
To achieve this goal, we propose a progressive multi-task learning framework that fully leverages the distinct focuses of each task on facial emotion features. Specifically, our method design can be summarized into three main aspects:
1) \textbf{Separate Training and Joint Training:} We first train each task model separately and then perform joint training based on the pre-trained models, fully utilizing the feature focus aspects of each task to improve the overall framework performance.
2) \textbf{Feature Fusion and Temporal Modeling:} We investigate effective strategies for fusing features extracted from each task-specific model and incorporate temporal feature modeling during the joint training phase, which further refines the performance of each task.
3) \textbf{Joint Training Strategy Optimization:} To identify the optimal joint training approach, we conduct a comprehensive strategy search, experimenting with various task combinations and training methodologies to further elevate the overall performance of each task.
According to the official results, our team achieves \textbf{first place} in the MTL challenge with a total score of 1.5286 (\emph{i.e.}, AU F-score 0.5580, Expression F-score 0.4286, CCC VA score 0.5420). Our code is publicly available \href{https://github.com/YenanLiu/ABAW7th}{here}.
\end{abstract}
    
\section{Introduction}
\label{sec:intro}
The objective of affective behavior analysis is to understand and interpret human emotions through various data sources such as facial expressions, speech, text, and physiological signals \cite{kollias20246th, kollias2020analysing, yin2023multi, kollias2021analysing, zhang2023multi, nguyen2023transformer, ritzhaupt2021meta, kollias2023abaw2, kollias2023abaw, kollias2021distribution, kollias2019expression, kollias2019deep, kollias2019face, zafeiriou2017aff}. 
By analyzing these affective cues, we can detect and quantify the emotional states of humans, thus enabling the development of systems that can interact with humans in a more natural and empathetic manner \cite{gervasi2023applications, vsumak2021sensors, filippini2020thermal, szaboova2020emotion, ren2023human}.

To facilitate this, the 7th Affective Behavior Analysis competition (ABAW7) \cite{kollias20247th} set two competition tracks, \emph{i.e.,} the Multi-task Learning (MTL) Challenge and the Compound Expression (CE) \cite{dong2024bi, he2022compound, she2021dive, wang2020suppressing} Challenge based on the Aff-Wild \cite{kollias2018aff} and C-EXPR-DB \cite{kollias2023multi} datasets.
Specifically, MTL actually involves three sub-tasks, \emph{i.e.} the action unit (AU) prediction, the expression recognition (EXPR) \cite{li2020deep, revina2021survey, wang2020suppressing, farzaneh2021facial, zhao2021learning}, and the valence-arousal (VA) estimation \cite{kollias2021affect, kollias2022abaw, kollias2023abaw, liu2023evaef, praveen2023audio}.
In this track, participants are encouraged to leverage the commonalities and differences across these tasks.
To improve the final performance of the model, we focus on three aspects: i) Enabling the model to learn general and powerful features that can be applied across tasks \cite{liu2023bavs, liu2023audio, zhang2024affectivebehaviouranalysisintegrating, qi2023emotiongesture, qi2023diverse}, enhancing the overall capability.
ii) Achieving knowledge transfer between tasks, so features effective for one task can benefit others.
iii) Fully utilizing the spatiotemporal features in the data to further boost model performance.

To be specific, we first integrate almost all the existing facial datasets and train a feature extractor in a self-supervised manner.
This extractor can be applied to all tasks to obtain strong and representative features.
To fully utilize the facial expression features captured by different models, we first train models for each task individually. 
Then we devise a feature fusion module to integrate the features extracted by these pre-trained models with the features of the current training model. 
Moreover, we employ a Temporal Convergence module to capture and integrate dynamic changes in expressions, fully exploiting the temporal information in the dataset.
Comprehensive evaluations conducted on the officially provided validation and test datasets the superiority of our method.

\section{Related Work}

Multi-task learning (MTL) has gained significant attention for its ability to leverage shared representations across related tasks, improving performance and generalization. In the task of Valence-Arousal (VA) estimation \cite{kollias2021affect, kollias2022abaw, kollias2023abaw}, Expression Recognition (EXPR) \cite{li2020deep, revina2021survey, wang2020suppressing}, and Action Unit (AU) detection \cite{li2021micro, tallec2022multi, belharbi2024guided}, more methods are being explored to achieve promising results.

VA estimation involves predicting continuous emotional dimensions and has been traditionally approached using single-task models. Recent MTL approaches \cite{Kollias_2019} have demonstrated that learning VA jointly with other tasks like ER and AU detection can enhance performance by capturing shared emotional cues. Their work utilized deep neural networks to jointly learn VA, ER, and AU, leading to improved generalizability.
Expression recognition has significantly benefited from MTL frameworks. 
Zhang \emph{et al.} \cite{zhang2015learningsocialrelationtraits, zhang2022multimodal} introduce an MTL approach that jointly learns ER with VA estimation and AU detection. 
Their model utilizes a shared feature extractor with task-specific heads, resulting in better accuracy and robustness across tasks. 
They highlighted the importance of capturing the interdependencies between expressions, valence, and arousal.
AU detection is another area where MTL has been effectively applied. 
Zhang \emph{et al.} \cite{zhang2024effective, zhang2024affectivebehaviouranalysisintegrating} propose a framework that simultaneously performs AU detection and ER, leveraging the correlations between AUs and expressions.

Recent advances in MTL have explored more sophisticated architectures and training strategies. 
For example, Akhtar \emph{et al.} \cite{akhtar2020deep} develop an attention-based MTL model that dynamically weights the importance of different tasks during training. 
This approach allows the model to focus on more relevant features for each task, leading to improved performance in VA, ER, and AU detection. 
Similarly, Zhao \emph{et al.} \cite{zhu2024cross} introduces a cross-task multi-branch vision transformer to ensure that predictions across tasks remain consistent, further enhancing the model's accuracy.
The development of comprehensive datasets like Aff-Wild2 \cite{kollias2018aff} and BP4D+  \emph{et al.} \cite{ZhangBP4D} has been crucial for advancing MTL in this domain. 
These datasets provide annotated data for VA, ER, and AU, enabling the training and evaluation of MTL models under diverse and challenging conditions.

\section{Method}
\label{sec:method}
 
\begin{figure*}[t]
  \centering
\includegraphics[width=1\linewidth]{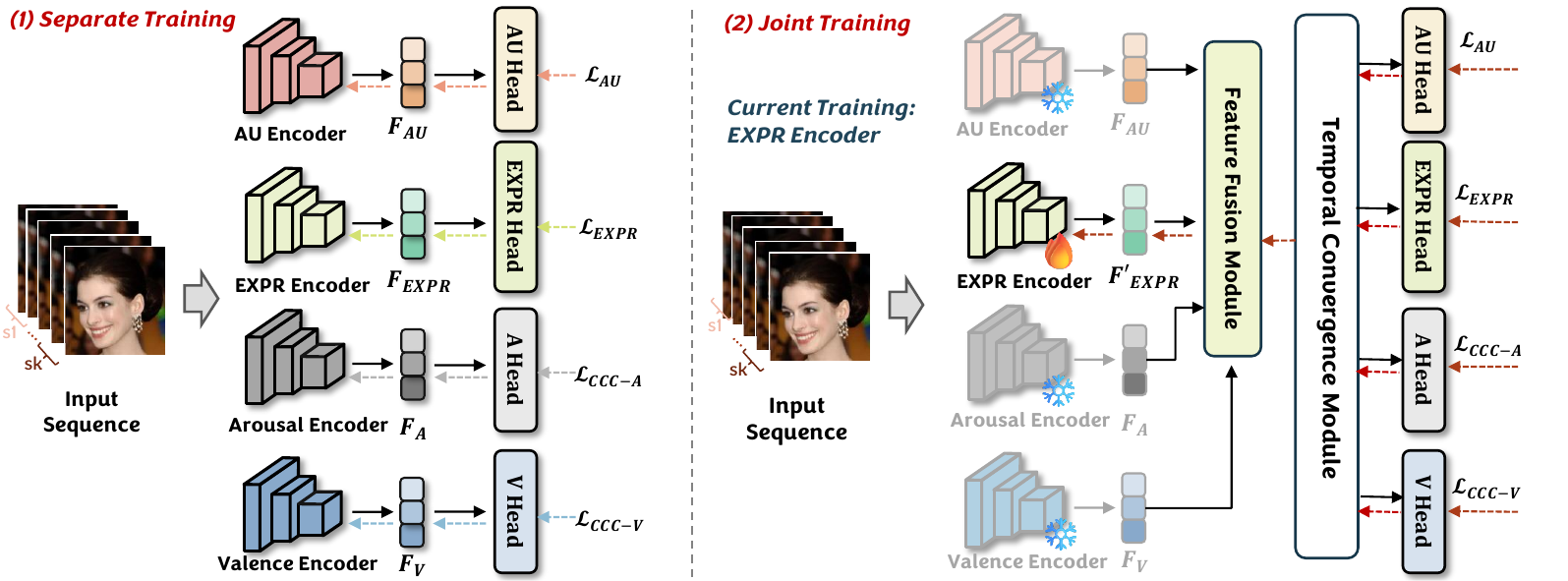}
\caption{\textbf{Illustration of our proposed framework for the MTL competition.} In the initial stage, we train models for each task separately. Once optimal performance is achieved, we begin joint training to enhance performance further. Taking EXPR model training as an example, features like $F_{AU}$, $F_{V}$, and $F_{A}$ from pre-trained encoders are fused with features from the current EXPR Encoder via the Feature Fusion Model. The fused features are then processed by the Temporal Convergence Module to capture temporal information. Finally, the features are sent to task-specific heads (AU, EXPR, and VA) for predictions. The final integration scheme for each subtask depends on the validation set performance, and the detailed analysis is provided in Sec. \ref{sec:exper} }
  \label{fig:pipeline}
\end{figure*}

\subsection{Multi-Task Learning}
\subsubsection{Method Overview.}
In the MTL track, as depicted in Fig. \ref{fig:pipeline}, we adopt a progressive learning approach, transitioning from single-task learning to multi-task joint learning. Initially, we construct classification or regression heads for the VA, ER, and AU tasks behind the MAE feature extractor and train them on the respective task's effective data (\emph{i.e.,} excluding data with VA labels of -5, and expression and AU labels of -1). 
Next, we explore the impact of joint training by combining various tasks. 
We employ the MAE as our feature extractor \cite{he2022masked, zhang2024affectivebehaviouranalysisintegrating, liu2024affectivebehaviouranalysisprogressive} for all sub-tasks. 
Additionally, we investigate the effect of integrating features from different sub-tasks into other tasks on model training performance. 
Specifically, we first extract the output features of the MAE for each task after individual training.
When training other tasks, we introduce a feature fusion module behind the MAE to combine the features of other tasks with those of the task being trained. 
We provide further results in the experimental section demonstrating the performance improvements from the progressive learning strategy.

\subsubsection{Feature Fused Module.}
We construct a feature fusion module $\mathcal{F}_f$ to fully utilize the expression information captured under different tasks. Specifically, our fusion module consists of two linear layers, an activation layer, and a Dropout layer. We input images  $I\in \{b \times s, 3, h, w\}$  into the pre-trained single-task model extractors to obtain features $F_{au}$, $F_{expr}$, and $F_{va}$, respectively. 
Here, $b$ and $s$ are the batch size and the sequence length of input images, and $h$ as well as $w$ indicate the image height and width respectively.
When training other tasks, we fuse these extracted features $F_{other}$ with the features $ F_{current}$ extracted by the model being trained. Then we feed the added feature into the subsequent modules of the model to attain the fused feature $F_m$, the process can be expressed as follows:

\begin{equation}
\label{eq:FeaFuse}
F_m = \mathcal{F}_f(F_{other} + F_{current}). 
\end{equation}

For example, when training the VA task, we add the pre-extracted $F_{au}$ features to the $F_{va'}$ features extracted by the current training model and then feed them into the fusion layer to obtain the fused features $F_m$. We analyze the promotion effect of this feature fusion method on different tasks in the experimental section.

\subsubsection{Temporal Convergence Module.}
To capture the expression dynamic changes over time, we construct the temporal convergence module.
Specifically, the model consists of two LSTM layers and a Leaky ReLU activation function. 
After obtaining the fused features $F_m \in \{b\times s, d\}$, we reshape it into $\{b, s, d\}$ where $b$ and $s$ represent the batch size and sequence length respectively, $d$ indicates the feature dimension.
The Temporal Convergence Module aggregates information across each time series $s$, resulting in more comprehensive expression features. 
It is worth noting that in the MTL track, not all tasks employ the Temporal Convergence Module.
It is only used in Arousal-Valence estimation and Expression Recognition tasks. 
Additionally, the performance of different tasks varies significantly with different window sizes and sequence lengths, which we will analyze in detail in the experimental section.
 
\subsection{Training Objectives}
\subsubsection{Objective for Multi-Task Learning.}
For the AU detection task, we utilize the Binary Cross Entropy (BCE) as the optimization objective, formulated as:

\begin{equation}
\mathcal{L}_{AU} = -\frac{1}{12} \sum_{j = 1}^{12} W_{a u_{j}}\left[y_{j} \log \hat{y}_{j}+\left(1-y_{j}\right) \log \left(1-\hat{y}_{j}\right)\right],
\label{eq:au}
\end{equation}
where  $y$ and $\hat{y}$ indicate the ground-truth and predicted labels, respectively.

For the EXPR task, we employ Cross Entropy (CE) Loss as our training objective, expressed as:
\begin{equation}
\mathcal{L}_{EXPR} = -\frac{1}{8} \sum_{j = 1}^{8} W_{exp{-}{j}} z_{j} \log \hat{z}_{j},
\label{eq:expr}
\end{equation}
where  $\hat{y}$ and $\hat{z}$ represent the predicted results for the action unit and expression category respectively, whereas $y$ and $z$ denote the ground truth values for the action unit and expression category.

In the VA estimation task, we leverage the consistency correlation coefficient as the model optimization function, defined as:
\begin{equation}
\operatorname{CCC}(\mathcal{X}, \hat{\mathcal{X}})=\frac{2 \rho_{\mathcal{X} \hat{\mathcal{X}}} \delta_{\mathcal{X}} \delta_{\hat{\mathcal{X}}}}{\delta_{\mathcal{X}}^{2}+\delta_{\hat{\mathcal{X}}}^{2}+\left(\mu_{\mathcal{X}}-\mu_{\hat{\mathcal{X}}}\right)^{2}},
\label{eq:ccc}
\end{equation}

\begin{equation}
\begin{split}
\mathcal{L}_ {\operatorname{VA}} =1-\operatorname{CCC}(\hat {v}_ {batch_{i}}, v_ {batch_{i}}) 
+ 1-\operatorname{CCC}(\hat{a}_ {batch_{i}}, a_{batch_{i}}).
\end{split}
\end{equation}
Here, $\hat{v}$ and $\hat{a}$ represent the predicted valence and arousal value. 
$\delta_{\mathcal{X}}$ and $\delta_{\hat{\mathcal{X}}}$ indicate the ground-truth sample set and the predicted sample set.
$\rho_{\mathcal{X} \hat{\mathcal{X}}}$ is the Pearson correlation coefficient between $\mathcal{X}$ and $\hat{\mathcal{X}}$, 
$\delta_{\mathcal{X}}$ and $\delta_{\hat{\mathcal{X}}}$ are the standard deviations of $\mathcal{X}$ and $\hat{\mathcal{X}}$, and $\mu_{\mathcal{X}}$, $\mu_{\hat{\mathcal{X}}}$ are the corresponding means. The numerator $2 \rho_{\mathcal{X} \hat{\mathcal{X}}} \delta_{\mathcal{X}} \delta_{\hat{\mathcal{X}}}$ represents the covariance between the $\delta_{\mathcal{X}}$ and $\delta_{\hat{\mathcal{X}}}$ sample sets.

When we adopt the joint-learning strategy to train the model, the training objective is formulated as follows:
\begin{equation}
\mathcal{L}_{Overall} = \lambda_{AU}\cdot\mathcal{L}_ {\operatorname{AU}} + \lambda_{EXPR}\cdot\mathcal{L}_ {\operatorname{EXPR}} + \lambda_{VA}\cdot\mathcal{L}_ {\operatorname{VA}},
\label{eq:overall}
\end{equation}
where $\lambda_{AU}$, $\lambda_{EXPR}$, and $\lambda_{VA}$ represent the weight of each sub-objective, respectively.
If a sub-task is not involved in joint training, its corresponding weight is set to zero.

\section{Experiments}
\label{sec:exper}

\subsection{Dataset}
For the MTL track, a total of 142,382 training images and 26,876 validation images from the Aff-wild \cite{kollias2018aff} dataset are provided by the organizers. After removing duplicate entries from the annotation files, the training and validation datasets consist of 141,431 and 26,666 images, respectively. After filtering out invalid data with AU labels of -1, the training and validation datasets contain 103,316 and 26,666 images, respectively. After filtering out invalid data with VA labels of -5, the training and validation datasets contain 103,917 and 26,666 images, respectively. After filtering out invalid data with EXPR labels of -1, the training and validation datasets contain 90,645 and 15,363 images, respectively. In this dataset, the values for Valence and Arousal range from [-1, 1]. There are 8 categories of Expression: Neutral, Anger, Disgust, Fear, Happiness, Sadness, Surprise, and Other. There are 12 types of Action Units: AU1, AU2, AU4, AU6, AU7, AU10, AU12, AU15, AU23, AU24, AU25, and AU26.

\subsubsection{Extra Data.}
In addition to the official data, we utilize the images from AffectNet \cite{mollahosseini2017affectnet}, CASIA-WebFace \cite{CASIA-Webface}, CelebA \cite{CelebA}, IMDB-WIKI \cite{IMDB-WIKI}, WebFace260M \cite{zhu2021webface260m}.
We train the initial feature extractor on these datasets using a facial reconstruction task. 
After that, we fine-tune the extractor on BPD4 \cite{ZhangBP4D} by predicting action units and then use this trained model as the initial feature extractor for all tasks.
 
\subsection{Metrics}
The performance measure $P$ is calculated by summing the following components: the mean Concordance Correlation Coefficient (CCC) for valence and arousal, the average F1 Score across all eight expression categories $F_{expr}$, and the average F1 Score across all 12 action units $F_{aus}$, mathematically formulated as follows:

\begin{equation}
    P = \left( \operatorname{CCC}_{arousal} + \operatorname{CCC}_{valence} \right) / 2 + F_{expr} + F_{aus}.
\end{equation}
 
\subsection{Implementation Details}
All training images are resized to 224 $\times$ 224 pixels. The MAE pertaining is conducted on 8 NVIDIA A30 GPUs with a batch size of 4096.
For training details of the multi-task learning track, all training configurations (including the batch size, optimizer, scheduler, and learning rate \emph{et al.}) can be found in our released code.
All experiments are conducted on 4 NVIDIA A100 GPUs.

\subsection{Experimental Results}

\subsubsection{Analysis on the Validation dataset.}
We evaluate our final approach to the MTL challenge on the official validation set. 
Since the CE challenge does not involve the validation set, we evaluate our method to the CE challenge on the validation set of RAF-DB-b and FUXI-EXPR.
Results are shown in Table \ref{tab:all}. 

\begin{table}[htp]
\vspace{-0.5em}
\centering
\caption{Quantitative results on the officially provided validation dataset.}
\label{tab:all}
\setlength{\tabcolsep}{0.5 em}
\renewcommand\arraystretch{1.1}
\scriptsize
\begin{tabular}{lllllll}
\toprule
Tracks        & Average CCC & CCC-V  & CCC-A  & $F_{expr}$ & $F_{aus}$ & P      \\ \hline
MTL           & 0.6533      & 0.6926 & 0.6139 & 0.5030	   & 0.6351	    & \textbf{1.7914} \\
Baseline(MTL) & \_          & \_     & \_     & \_         & \_        & 0.32   \\ \bottomrule
\end{tabular}
\end{table}

\subsubsection{Impact of Feature Fusion Module.}
To evaluate the performance of each task model after employing the feature fusion module, we incorporate the features from other task models when training the current task model.
Specifically, we first attain the optimist model for each task and save the features extracted by their feature extractors. 
Then, we fuse features from other models and the current training model.
The results are shown in Table \ref{tab:ffm}.


\begin{table}
\centering
\caption{\textbf{Impact of the Feature Fusion Module.} All experiments are conducted based on separately trained models. Here, $Per.$ represents the performance of each model on its respective task.}
\label{tab:ffm}
\setlength{\tabcolsep}{1.1 em}
\renewcommand\arraystretch{0.8}
\scriptsize
\begin{tabular}{c|cccc|l}
\toprule
Sub-Tasks & w/$F_{AU}$                 & w/$F_{EXPR}$               & w/$F_{V}$            & w/$F_{A}$            & $Per.$          \\ \midrule
AU        &                            &                            &                            &                            & \textbf{0.5001} \\ 
          &                            & \ding{51} &                            &                            & 0.4826          \\
          &                            &                            & \ding{51} &                            & 0.4988          \\
          &                            &                            &                            & \ding{51} & 0.4974          \\ \midrule
EXPR      &                            &                            &                            &                            & 0.4012          \\
          & \ding{51} &                            &                            &                            & \textbf{0.4142} \\
          &                            &                            & \ding{51} &                            & 0.4105          \\
          &                            &                            &                            & \ding{51} & 0.3941          \\
          & \ding{51} &                            & \ding{51} &                            & 0.4207          \\ \midrule
Valence   &                            &                            &                            &                            & 0.5606          \\
          & \ding{51} &                            &                            &                            & 0.5721          \\
          &                            & \ding{51} &                            &                            & 0.5208          \\
          & \ding{51} &                            &                            & \ding{51} & \textbf{0.5780} \\ \midrule
Arousal   &                            &                            &                            &                            & 0.4026          \\
          & \ding{51} &                            &                            &                            & 0.4142          \\
          &                            & \ding{51} &                            &                            & 0.3987          \\
          &                            &                            & \ding{51} &                            & 0.4297          \\
          & \ding{51} &                            & \ding{51} &                            & \textbf{0.4581} \\ \bottomrule
\end{tabular}
\end{table}

As suggested in Table \ref{tab:ffm}, combining features from other task models via the feature fusion module does not universally enhance the performance of all task models. The degree of performance improvement varies significantly depending on the fused features. Specifically, for the AU task model, the best performance (F-score of 0.5001) is achieved without using the feature fusion module, and introducing features from other task models decreases performance by 0.28\% to 1.75\%. 
Conversely, for the EXPR task model, fusing features from the AU task model improves performance by 1.4\%. 
For the Valence task model, the best results come from incorporating features from both AU and Arousal models, resulting in a 1.74\% improvement. 
For the Arousal task, integrating features from both the AU and Valence models leads to a significant performance increase of 5.69\%.

These variations can be attributed to the training quality of the different models. As indicated in Table \ref{tab:ffm}, the F-score values of the Expression and Arousal prediction models are slightly lower, at 0.4012 and 0.4026, respectively.
This suggests that the quality of features extracted by these models is lower. Therefore, incorporating features from these two models is likely to decrease the performance of the current model.

\subsubsection{Impact of Temporal Convergence Module.}
To evaluate the impact of introducing the Temporal Convergence Module on different tasks, we incorporate this module into each task model and analyze the effects of varying sequence length and window size on the final results. 
As suggested in Table \ref{tab:tcm}, the performance of AU and EXPR task models is not enhanced.
However, the module achieves significant improvements in the Valence and Arousal estimation models.
Specifically, the CCC value attains 2.36\% and 3.45\% improvement respectively on the two tasks compared with training on single frames.

\begin{table} 
\caption{\textbf{Impact of the Temporal Convergence Module.} All experiments are conducted based on independently trained models. Here, $Per.$ represents the performance of each model on its respective task, Tem. is whether training in a sequence manner and $S$ as well as $W$ indicate the sequence length and window size, respectively.}
\centering
\label{tab:tcm}
\setlength{\tabcolsep}{2.4 em}
\renewcommand\arraystretch{1.0}
\scriptsize
\begin{tabular}{c|cccl}
\toprule
Sub-Tasks             & Tem. & S & W  & $Per.$          \\ \midrule
\multirow{3}{*}{AU}   & \ding{51}                   &                 &             & \textbf{0.5001} \\
                      &                      & 5               & 5           & 0.4912          \\
                      &                      & 10              & 10          & 0.4599          \\ \midrule
\multirow{3}{*}{EXPR} & \ding{51}                 &                 &             & \textbf{0.4012}          \\
                      &                      & 5               & 5           &  0.3956               \\
                      &                      & 10              & 10          &  0.3876               \\ \midrule
\multirow{6}{*}{V}    & \ding{51}                  &                 &             & 0.5606          \\
                      &                      & 5               & 5           & 0.5705          \\
                      &                      & 10              & 10          & 0.5774          \\
                      &                      & 15              & 15          & 0.5831          \\
                      &                      & 20              & 20          & \textbf{0.5842} \\
                      &                      & 20              & 15          & 0.5806          \\ \midrule
\multirow{7}{*}{A}    & \ding{51}                  &                 &             & 0.4012          \\
                      &                      & 5               & 5           & 0.4148          \\
                      &                      & 10              & 10          & 0.4248          \\
                      &                      & 15              & 15          & 0.4231          \\
                      &                      & 20              & 20          & 0.4304          \\
                      &                      & 20              & 15          & 0.4276          \\
                      &                      & 30              & 30          & \textbf{0.4357} \\ \bottomrule
\end{tabular}
\end{table}

Moreover, we also experiment with different sequence lengths and window sizes on the Valence and Arousal estimation models. 
We found that adjusting the window size does not significantly impact performance. However, changes in sequence length have a more considerable effect. 
The Valence estimation model achieves the best performance with a sequence length of 20. 
For the Arousal prediction model, the best performance is achieved with a sequence length of 30, reaching an F-score of 0.4357.

\subsubsection{Impact of Joint Training strategy.}

To achieve the optimal results for each task, we introduce the joint training strategy and explore the improvement effects of joint training on each task's results. 
As shown in Table \ref{tab:joint}, the performance of all task models is improved to varying degrees after training with other tasks. 
We also explore how combining the Joint Training strategy with the Temporal Convergence Module and Feature Fusion Module achieves the best performance across all tasks. 
Note that, we only present some most competitive experimental results in Table \ref{tab:joint} for each task.
For the AU task model, training with the Valence estimation model achieves the most optimal result.

For the EXPR task model, the highest F-score of 0.5030 is achieved by fusing features from the AU and Valence models and jointly training with the AU, EXPR, and Valence estimation tasks.
For the Valence estimation task, the best performance, with a CCC value of 0.6926, is obtained by combining features from the AU and Valence models and training under the Valence, Arousal estimation, and AU prediction tasks. 
For the Arousal estimation task, the optimal result is achieved by incorporating features from the AU and Valence models and jointly training with the AU, Valence, and Arousal estimation tasks.


\begin{table}[htp]
\caption{\textbf{Impact of the Joint Training.} We combine all strategies together to explore the optimist combination. Here, $Per.$ represents the performance of each model on its respective task, Tem. is whether training in a sequence manner, and $S$ as well as $W$ indicate the sequence length and window size, respectively.}
\label{tab:joint}
\centering
\setlength{\tabcolsep}{0.6 em}
\renewcommand\arraystretch{1.2}
\scriptsize
\begin{tabular}{c|ccc|cccc|cc|l}
\hline
\multirow{2}{*}{}     & \multicolumn{3}{c|}{Featrue Fusion Setting} & \multicolumn{4}{c|}{Joint Training Task} & \multicolumn{2}{c|}{Tem.} & \multirow{2}{*}{Per.} \\ \cline{2-10}
                      & w/$F_{AU}$   & w/$F_{EXPR}$   & w/$F_{V}$   & AU       & EXPR      & V       & A       & S           & W           &                       \\ \hline
AU                    &              &                &             &          &           & \ding{51}      &         & \_          & \_          & \textbf{0.6351}       \\ \hline
\multirow{4}{*}{EXPR} & \ding{51}           & \ding{51}             & \ding{51}          & \ding{51}      & \ding{51}        & \ding{51}      & \ding{51}      & 5           & 5           & 0.4690                \\
                      & \ding{51}           &                & \ding{51}          & \ding{51}       & \ding{51}        & \ding{51}      & \ding{51}      & 5           & 5           & 0.4722                \\
                      & \ding{51}           &                & \ding{51}          & \ding{51}       & \ding{51}        & \ding{51}      &         & 5           & 5           & \textbf{0.5030}       \\
                      & \ding{51}          &                &             & \ding{51}       & \ding{51}        & \ding{51}      &         & 5           & 5           & 0.4586                \\ \hline
\multirow{6}{*}{V}    & \ding{51}           &                &             & \ding{51}       & \ding{51}        & \ding{51}      & \ding{51}      & 10          & 10          & 0.6271                \\
                      & \ding{51}           &                &             & \ding{51}       &           & \ding{51}      & \ding{51}      & 10          & 10          & 0.6354                \\
                      &              &                & \ding{51}          & \ding{51}       & \ding{51}        & \ding{51}      & \ding{51}      & 10          & 10          & 0.6041                \\
                      & \ding{51}           &                & \ding{51}          & \ding{51}       &           & \ding{51}      & \ding{51}      & 10          & 10          & 0.6456                \\
                      & \ding{51}           &                & \ding{51}          & \ding{51}       &           & \ding{51}      & \ding{51}      & 15          & 15          & 0.6487                \\
                      & \ding{51}           &                & \ding{51}          & \ding{51}       &           & \ding{51}      & \ding{51}     & 20          & 20          & \textbf{0.6926}       \\ \hline
\multirow{7}{*}{A}    & \ding{51}           &                &             & \ding{51}       & \ding{51}        & \ding{51}      & \ding{51}      & 10          & 10          & 0.5633                \\
                      & \ding{51}           &                &             & \ding{51}       &           & \ding{51}      & \ding{51}      & 10          & 10          & 0.5633                \\
                      &              &                & \ding{51}          & \ding{51}       & \ding{51}        & \ding{51}      & \ding{51}      & 10          & 10          & 0.5683                \\
                      & \ding{51}           &                & \ding{51}          & \ding{51}       &           & \ding{51}      & \ding{51}      & 10          & 10          & 0.5725                \\
                      & \ding{51}           &                & \ding{51}          & \ding{51}       &           & \ding{51}      & \ding{51}     & 15          & 15          & 0.6043                \\
                      & \ding{51}           &                & \ding{51}         & \ding{51}       &           & \ding{51}      & \ding{51}      & 20          & 20          & 0.5981                \\
                      & \ding{51}          &                & \ding{51}          & \ding{51}       &           & \ding{51}      & \ding{51}      & 30          & 30          & \textbf{0.6139}       \\ \hline
\end{tabular}
\end{table}

\subsection{Evaluation on the Test Dataset}
In Table \ref{tab:final}, we show the final evaluations on the official test dataset.
Our team (\emph{i.e.} Netease Fuxi AI Lab) attains a first place in the Multi-Task Learning competition.
Overall, our framework outperforms the second-place team (\emph{i.e.,} HSEomotion) by approximately 28.14\%. 
Specifically, our method demonstrates exceptional performance in the VA Estimation task, surpassing the second-place team by nearly 14\%. 
Additionally, in the AU Detection and Expression Recognition tasks, our method outperforms the second-place team by 4.61\% and 9.89\%, respectively. 
This fully proves the effectiveness of our solution. 
As suggested by the results of all teams, the F1 score for the Expression Recognition task was relatively lower compared to the other two tracks. 
We attribute this to the limited amount of training data.
The amount of the training data for Expression Recognition is one-third less than that for the other two tracks, while the amount of validation and test data is roughly the same across all three tracks.

\begin{table}[]
\caption{\textbf{There are the final results on the official test dataset.} Our team ( \textbf{\emph{i.e.} Netease Fuxi AI Lab}) attains the best results on MTL, outperforming the second place by 28.14\% in the overall metric. Moreover, our method also shows outstanding performance in the AU, EXPR, and VA tasks.}
\label{tab:final}
\centering
\setlength{\tabcolsep}{0.2 em}
\renewcommand\arraystretch{1.2}
\scriptsize
\begin{tabular}{c|cccc}
\hline
Teams                      & Overall Metric & AUs F1 Score & EXPR F1 Score & CCC VA \\ \hline
baseline \cite{kollias20247th}                  & 0.3400         & 0.1200       & 0.1007               & 0.1193       \\ \hline
SML                      & 0.8692         & 0.4046       & 0.1938               & 0.4074       \\
AIWELL-UOC  \cite{cabacasmaso2024enhancingfacialexpressionrecognition}               & 1.1145         & 0.4663       & 0.2772               & 0.3783       \\
SCU ACers \cite{li2024affectivebehavioranalysisusing}                 & 1.1640         & 0.4879       & 0.3018               & 0.3743       \\
HFUT-MAC1  \cite{shen2024facialaffectrecognitionbased}                & 1.1777         & 0.4997       & 0.2997               & 0.3783       \\
HSEmotion  \cite{savchenko2024hsemotionteam7thabaw}                & 1.2472         & 0.5119       & 0.3279               & 0.4074       \\
\textbf{Netease Fuxi AI Lab} \cite{liu2024affectivebehaviouranalysisprogressive} & \textbf{1.5286 }        & \textbf{0.5580}       & \textbf{0.4286  }             & \textbf{0.5420}       \\ \hline
\end{tabular}
\end{table}
\vspace{-1.0em}
\section{Conclusion}
\label{sec:Conclu}

In this paper, we present our methods for the 7th ABAW Multi-task Learning Challenge.
Based on our re-collected large-scale facial dataset, we train a robust facial feature extractor. Meanwhile, based on the pretrained feature extractor, we finetune it on downstream task datasets thus obtaining the high-quality facial expression feature. 
We devise the feature fusion model to fuse the features extracted from other affective analysis tasks with features extracted from the current model, which improves the model performance for EXPR, Valence, and Arousal estimation models.
Moreover, our temporal convergence model captures the expression dynamic changes over sequence frames, enhancing the model performance for  Valence and Arousal estimation models.
Comprehensive experiments demonstrate the superiority of our method compared to other solutions.

\vspace{1.0em}
\noindent\textbf{Acknowledgements.} This research is funded in part by the National Key R\&D Program of China (No. 2022YFF09022303 to Lincheng Li), ARC-Discovery grant (DP220100800 to XY), and ARC-DECRAgrant (DE230100477 to XY). The first author is funded by the CSC scholarship and Data61-topup scholarship.
 
\bibliographystyle{splncs04}
\bibliography{main}
\end{document}